\documentclass[11pt,a4paper,twocolumn]{article}

\usepackage[T1]{fontenc}
\usepackage[utf8]{inputenc}
\usepackage{times}
\usepackage[top=2.5cm,bottom=2.5cm,left=2.5cm,right=2.5cm,columnsep=0.6cm]{geometry}
\usepackage{amsmath,amssymb}
\usepackage{graphicx}
\usepackage{booktabs}
\usepackage{tabularx}
\usepackage{array}
\usepackage{caption}
\usepackage{titlesec}
\usepackage[numbers,sort&compress]{natbib}
\usepackage{xcolor}
\definecolor{linkblue}{rgb}{0,0,0.5}
\usepackage[colorlinks=true,citecolor=linkblue,linkcolor=linkblue,urlcolor=linkblue]{hyperref}
\usepackage{tikz}
\usetikzlibrary{positioning,arrows.meta,fit,calc}

\titleformat{\section}{\normalfont\large\scshape}{\thesection}{1em}{}
\titleformat{\subsection}{\normalfont\normalsize\scshape}{\thesubsection}{1em}{}
\titlespacing*{\section}{0pt}{2.2ex plus 0.5ex}{1.2ex plus 0.2ex}
\titlespacing*{\subsection}{0pt}{1.8ex plus 0.4ex}{0.9ex plus 0.2ex}
\titlespacing*{\paragraph}{0pt}{1ex plus 0.3ex}{0.6em}
\newcolumntype{L}{>{\raggedright\arraybackslash}X}

\makeatletter
\renewcommand{\maketitle}{\twocolumn[{\centering
  {\Large\bfseries \@title\par}\vspace{1.2em}
  {\large\bfseries \@author\par}\vspace{1.6em}
  {\large\bfseries\scshape Abstract\par}\vspace{0.5em}}
  {\begin{list}{}{\setlength{\leftmargin}{1.3cm}\setlength{\rightmargin}{1.3cm}\setlength{\topsep}{0pt}\setlength{\parsep}{0.45em}\setlength{\itemsep}{0pt}\setlength{\listparindent}{0pt}\setlength{\itemindent}{0pt}\setlength{\partopsep}{0pt}}
   \item\relax\@abstracttext
   \end{list}\par}\vspace{0.9em}
  {\noindent\textbf{\textit{K}eywords}\quad \@keywordstext\par}\vspace{1.4em}]}
\newcommand{\abstracttext}[1]{\gdef\@abstracttext{#1}}
\newcommand{\keywordstext}[1]{\gdef\@keywordstext{#1}}
\makeatother

\title{Occupancy Modelling of Biased Landslide Inventories:\\
When Does Separating Occurrence from Recording Improve on Neutralisation?}
\author{\begin{tabular}{c@{\hspace{3em}}c}
Abdirashid Omar & Abdikadir Ugaas\\[0.2em]
{\normalfont\normalsize\ttfamily rashiidmatan@gmail.com} & {\normalfont\normalsize\ttfamily ugas@chungbuk.ac.kr}
\end{tabular}}

\abstracttext{%
Landslide susceptibility models are trained on inventories that record where landslides were seen, not where
they occurred. The established remedy fits bias-describing variables and neutralises them when mapping. We adapt
occupancy--detection modelling from ecology, treating several inventories as imperfect observers of one event
and separating occurrence from recording, and ask when this improves on neutralisation.

We use near-complete inventories of the 2018 Hokkaido Eastern Iburi, 2015 Gorkha and 2010 Haiti earthquakes with
simulated reporting bias, and three real Gorkha inventories. Compared with standard practice, the occupancy model
reduces road bias in 24 of 25 paired draws and raises AUC by 0.08--0.14 on Gorkha and 0.03 on Haiti, but only
0.01--0.02 on Iburi. Its median predicted-to-true landslide rate is 0.99 (range 0.44--1.89), against 0.24--0.30
for the baselines. Rescaling the baselines with the same anchor area corrects their median but leaves larger
errors (range 0.20--4.34). Cell-level probabilities remain too extreme (calibration slope 0.59). Compared with the
best neutralised model, AUC is higher on Iburi (by 0.03--0.07) and lower on Gorkha and Haiti (by 0.01--0.05).
Transplant experiments tie the Iburi advantage to the structure of its true landslide pattern, not to its terrain
or road network. In all eight settings with a low interaction gap, computable from biased records, neutralisation was at least as
accurate, including a prespecified test on Haiti; a high gap did not guarantee an advantage. The real Gorkha inventories are slope-biased, not road-biased, and there the model brings no meaningful
gain. Report-based catalogues remain the main untested case.
}

\keywordstext{Landslide susceptibility mapping $\cdot$ Landslide inventory bias $\cdot$ Occupancy--detection modelling $\cdot$ Imperfect detection $\cdot$ Bias neutralisation $\cdot$ Probability calibration $\cdot$ Earthquake-triggered landslides}

\begin{document}
\maketitle


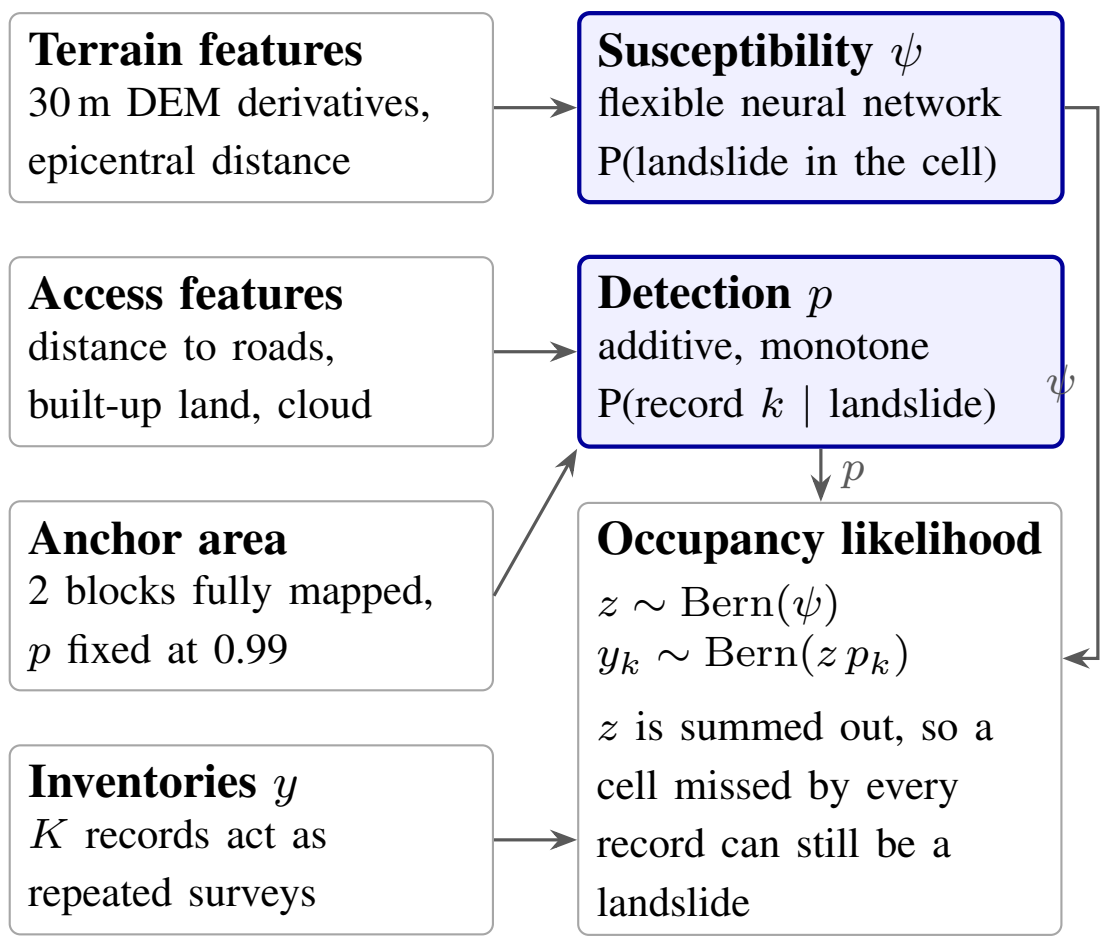
\begin{figure}[t]
\centering
\begin{tikzpicture}[
  font=\footnotesize,
  box/.style={draw=gray!70, rounded corners=2pt, align=left, inner sep=3.5pt, text width=2.95cm, minimum height=1.25cm},
  key/.style={box, draw=blue!60!black, fill=blue!6, line width=0.8pt},
  arr/.style={-{Stealth[length=1.8mm]}, gray!70!black, line width=0.6pt},
  node distance=0.35cm and 0.55cm]
  \node[box] (terrain) {\textbf{Terrain features}\\ \scriptsize 30\,m DEM derivatives,\\ \scriptsize epicentral distance};
  \node[key, right=of terrain] (psi) {\textbf{Susceptibility $\psi$}\\ \scriptsize flexible neural network\\ \scriptsize P(landslide in the cell)};
  \node[box, below=of terrain] (access) {\textbf{Access features}\\ \scriptsize distance to roads,\\ \scriptsize built-up land, cloud};
  \node[key, right=of access] (det) {\textbf{Detection $p$}\\ \scriptsize additive, monotone\\ \scriptsize P(record $k$ $\mid$ landslide)};
  \node[box, below=of access] (anchor) {\textbf{Anchor area}\\ \scriptsize 2 blocks fully mapped,\\ \scriptsize $p$ fixed at 0.99};
  \node[box, below=of anchor] (inv) {\textbf{Inventories $y$}\\ \scriptsize $K$ records act as\\ \scriptsize repeated surveys};
  \node[box, right=of inv, yshift=0.8cm, minimum height=2.85cm] (lik) {\textbf{Occupancy likelihood}\\[2pt] \scriptsize $z \sim \mathrm{Bern}(\psi)$\\ \scriptsize $y_k \sim \mathrm{Bern}(z\,p_k)$\\[3pt] \scriptsize $z$ is summed out, so a cell missed by every record can still be a landslide};
  \draw[arr] (terrain) -- (psi);
  \draw[arr] (access) -- (det);
  \draw[arr] (anchor.east) -- (det.south west);
  \draw[arr] (det) -- node[right]{\scriptsize $p$} (lik.north -| det);
  \draw[arr] (psi.east) -- ++(0.22,0) |- node[pos=0.25,left]{\scriptsize $\psi$} ($(lik.east)+(0,0.4)$);
  \draw[arr] (inv) -- (inv -| lik.west);
\end{tikzpicture}
\caption{Model structure. The susceptibility head sees only terrain and the detection head sees only access. Both enter the occupancy
likelihood, fitted to several inventories at once, with the anchor area fixing the scale of detection. Only $\psi$ is mapped.}
\label{fig:model}
\end{figure}

\section{Introduction}

Data-driven landslide susceptibility maps depend on the inventories they learn from \citep{reichenbach2018}. An
inventory is a record of observation: a landslide enters it when someone reports, surveys or maps it. Reporting is
easier near roads and towns and under clear skies, so inventories are spatially incomplete in systematic ways
\citep{steger2017,bornaetxea2022}. A classifier trained on such records learns the observation process together with
the hazard. The problem matters because landslides do not cluster near roads in the way reports do. In all three
earthquakes studied here, cells with mapped landslides lie farther from public roads than other cells (median 0.50
vs 0.25\,km in Iburi, 0.66 vs 0.33\,km in Gorkha, 0.81 vs 0.62\,km in Haiti), so road-biased records under-sample
terrain where landslides are common.

Dropping bias-related predictors confounds the remaining effects \citep{steger2017}. The established alternative
is to include them during fitting and remove their influence when mapping \citep{steger2021}. This neutralisation
needs only one inventory, but it assumes that the bias variables can be separated from the hazard signal inside a
single model. Ecology has long faced the same problem, since species are recorded where observers look. Occupancy
models address it structurally, with one process for true presence and another for detection, identified through
repeated surveys \citep{mackenzie2002}. Several independently produced inventories of one event could play the role
of those surveys, but this requires more data than neutralisation, and whether it is worth the cost is an open
question.

This paper therefore asks when treating inventories as imperfect observers improves on neutralisation. We make three
contributions:
\begin{enumerate}
  \item \textbf{A model.} We adapt deep occupancy--detection modelling to landslide inventories, using separate
        inventories of one event as repeated surveys, sign-constrained detection, and a small exhaustively mapped
        anchor area.
  \item \textbf{An evaluation against a known reference.} On real terrain, roads and near-complete inventories from
        three earthquakes with simulated reporting bias, the model reduces road bias relative to standard practice
        and estimates the overall landslide rate more stably than baselines, including baselines rescaled with the
        same anchor. Its ranking advantage over neutralisation is confined to one earthquake. On three real Gorkha
        inventories, whose bias follows slope rather than roads, it brings no meaningful gain.
  \item \textbf{An explanation and a selection rule.} Transplant experiments link the ranking advantage to the
        structure of the true landslide pattern rather than to terrain or roads. From this we derive a rule that can
        be computed from biased records. In our tests a low value always meant that the simpler neutralisation
        sufficed, including a prediction recorded before testing on a third earthquake. A high value is necessary
        but not sufficient for an advantage.
\end{enumerate}

\section{Related work}

\paragraph{Inventory bias in susceptibility mapping.}
Statistical susceptibility models are reviewed by \citet{reichenbach2018}. Three lines of work show that inventory
bias changes model results. \citet{steger2017} introduced synthetic biases and showed that excluding bias-describing
predictors confounds the remaining effects, while mixed-effects models were least affected. \citet{stanley2017}
truncated a synthetic inventory to mimic road and population reporting biases and found that the effects should not
be ignored. In Bavaria, gaps in the inventory led to underestimated hazard in poorly mapped areas
\citep{buchauer2026}. How bias arises depends on how an inventory was made. Field, report and newspaper inventories
under-sample terrain poorly visible from roads \citep{bornaetxea2022}, whereas excess landslide density near roads
appeared in only 3 of 12 inventories examined by \citet{malamud2022}, who attributed it to mapping remit. As remedies,
\citet{steger2021} and \citet{lima2021} account for data-collection effects during fitting and remove them for
prediction, the latter at national scale. \citet{steger2021} enter bias-describing variables as random effects in a
generalised additive mixed model and zero them for prediction. In their South Tyrol case this bias-adjusted model had
limited success (median AUROC 0.59--0.61), which they attribute partly to test data that share the bias. These remedies
use one inventory and one model of the combined process, which is the assumption we examine.

\paragraph{Negative-sample selection.}
A large literature treats bias as a problem of choosing non-landslide samples, for example positive--unlabelled
learning with adaptive sampling \citep{fang2021} or with a contrastive loss \citep{ouyang2024}. These methods address
unknown negatives, but not spatially structured under-reporting of positives.

\paragraph{Inventory comparison.}
Independent inventories of the Gorkha earthquake differ substantially in completeness and delineation
\citep{pokharel2021}, and co-seismic inventories worldwide vary widely in quality and completeness \citep{tanyas2017}.
The USGS ground-failure repository collects them in one format \citep{schmitt2017}, so multiple inventories of one
event are available, but they have rarely been used jointly in model fitting.

\paragraph{Occupancy models.}
Ecological occupancy models separate presence from imperfect detection using repeated surveys \citep{mackenzie2002},
and have been combined with neural networks \citep{joseph2020,seo2021,benerradi2026}. False-positive extensions allow
observers to record absent species \citep{royle2006} and become well identified when some detections are known to be
certain \citep{miller2011}. To our knowledge, occupancy modelling has not been applied to landslide inventories.

\section{Method}

\paragraph{Occupancy--detection likelihood.}
For grid cell $i$, let $z_i = 1$ if a landslide occurred, with probability $\psi_i$ given by a susceptibility network
on terrain features. Inventory $k$ records a landslide in cell $i$ with probability $p_{ik}$ given by a detection
network on observation features, and only where it surveyed ($m_{ik} = 1$). With records $y_{ik}$, the likelihood of a
cell marginalises over the unknown $z_i$:
\begin{equation}
  \begin{split}
  L_i ={}& \psi_i \prod_k \left[p_{ik}^{\,y_{ik}} (1-p_{ik})^{1-y_{ik}}\right]^{m_{ik}} \\
        &+ (1-\psi_i)\,\mathbb{1}\!\left[\textstyle\sum_k m_{ik} y_{ik} = 0\right].
  \end{split}
  \label{eq:lik}
\end{equation}
The two networks are trained jointly by minimising the mean negative log-likelihood. Only $\psi$ is used for mapping
(Fig.~\ref{fig:model}).

\paragraph{Monotone detection.}
Detection is additive and sign-constrained, in the spirit of shape-constrained additive models \citep{pya2015}: being
farther from a road, or under more cloud, can only lower the recording probability, and more built-up area can only
raise it. Each standardised feature $x_{ij}$ enters through a sum of increasing sigmoids with non-negative weights,
multiplied by its known sign $s_j$:
\begin{equation}
  \begin{split}
  \operatorname{logit} p_{ik} ={}& b_k + \sum_j s_j \sum_m \operatorname{softplus}(a_{kjm}) \\
  &\times \sigma\!\left(2\operatorname{softplus}(c_{kjm})\,(x_{ij} - t_m)\right),
  \end{split}
  \label{eq:det}
\end{equation}
where $\sigma$ is the logistic function and $t_m$ are eight fixed knots on $[-2.5, 2.5]$. Features whose effect
direction is unknown (terrain visibility for satellite inventories) enter twice, once with each sign, giving an
unconstrained additive shape.

\paragraph{Anchor area.}
Two training blocks of $25 \times 25$ cells (3\,km at 120\,m, 5.25\,km at 210\,m; about 2\% of the area) are treated as
exhaustively mapped. They enter the likelihood as one further survey, covering only those blocks, with detection fixed
at 0.99. This sets the scale of detection, and hence the overall landslide rate.

\paragraph{Assumptions.}
Identifying $\psi$ and $p$ separately rests on four assumptions: (i) all inventories observe the same $z$; (ii)
inventories are independent given $z$ and the features; (iii) there are no false positives, since Eq.~(\ref{eq:lik})
gives a recorded cell zero probability under $z_i = 0$; and (iv) an exclusion restriction, whereby access features
affect detection and terrain features affect occurrence. Assumption (iv) matters more than the number of inventories
suggests. In an early unconstrained version on Iburi, three inventories gained little over one merged inventory when
susceptibility used terrain only (AUC 0.75 vs 0.74), and more when it could also use road distance (0.74 vs 0.71).
Assumption (ii) is tested in Sect.~\ref{sec:robust}, and assumption (iii) fails for real inventories
(Sect.~\ref{sec:real}).

\paragraph{Training.}
We use full-batch AdamW \citep{loshchilov2019} in PyTorch \citep{paszke2019}, with learning rate $5 \times 10^{-3}$,
weight decay $10^{-4}$ and at most 3000 steps. Training stops early on a held-out validation block using observed
records only. Each model is the mean of three restarts. Networks are multilayer perceptrons (MLPs) with two hidden
layers of 64 SiLU units.

\paragraph{Extensions.}
Three extensions are tested in Sects.~\ref{sec:real} and \ref{sec:robust}. A short-range road term lets road distance
$d_i$ enter susceptibility only through a non-negative effect with a range of 50--300\,m, so that reporting bias acting
over kilometres cannot be absorbed:
\begin{equation}
  \begin{split}
  \operatorname{logit} \psi_i ={}& f(\text{terrain}_i) + \operatorname{softplus}(\alpha)\, e^{-d_i/\lambda},\\
  & \lambda \in [0.05, 0.30]\ \text{km}.
  \end{split}
  \label{eq:roadcut}
\end{equation}
A constant false-positive term gives each inventory a learned rate of recording cells with $z_i = 0$. A spillover term
models such records as arising mainly from neighbouring landslides (positional mismatch), with $q_i$ the probability
that at least one of the 8 neighbours $N(i)$ has a landslide:
\begin{equation}
  \begin{split}
  P(y_{ik}=1 \mid z_i=0) &= \sigma(\gamma_k)\, q_i + \sigma(\delta_k),\\
  q_i &= 1 - \prod_{j \in N(i)} (1-\psi_j).
  \end{split}
  \label{eq:spill}
\end{equation}

\paragraph{Selection rule.}
The interaction gap is the test-block AUC of a flexible gradient-boosting model
\citep{friedman2001,pedregosa2011} (300 iterations, learning rate 0.05) minus that of an additive spline model
\citep{eilers1996} (cubic splines, 6 knots), both on terrain features only. A gap above 0.03 predicts that the
occupancy model will outperform the best neutralised model, and a gap below 0.015 predicts that neutralisation will be
as good or better. Between these values the rule makes no prediction. The thresholds were set after measuring Iburi
(0.058) and Gorkha (0.000) on their reference inventories and were fixed before Haiti was analysed. Because
practitioners do not have a reference inventory, we also compute the gap from the union of biased records alone, for
both fitting and scoring.

\section{Data and experimental design}

\begin{table*}[t]
\centering
\caption{Study sites. Haiti was analysed only after the selection rule was fixed (Sect.~\ref{sec:haiti}).}
\label{tab:data}
\small
\begin{tabularx}{\textwidth}{@{}p{2.6cm}LLL@{}}
\toprule
 & 2018 Hokkaido Eastern Iburi & 2015 Gorkha (Sindhupalchok window) & 2010 Haiti \\
\midrule
Reference inventory & \citet{zhang2019}, 5,625 polygons & \citet{roback2018,zekkos2017}, source areas, 12,246 polygons & \citet{harp2016}, 23,567 polygons (21,666 in window) \\
Area and mask & convex hull of the inventory & mapped extent minus obscured areas, 27.65--28.05$^\circ$N, 85.40--86.10$^\circ$E & convex hull of the inventory, 18.18--18.50$^\circ$N, 72.98--72.17$^\circ$W \\
Grid & 120\,m, 56,399 cells, 16.3\% landslide & 210\,m, 62,652 cells, 10.4\% landslide & 210\,m, 63,995 cells, 10.4\% landslide \\
Epicentres & mainshock & mainshock and 12 May $M_w$\,7.3 aftershock & mainshock \\
Real independent inventories & -- & \citet{gnyawali2017}, \citet{valagussa2021}, \citet{zhangj2016} & -- \\
\bottomrule
\end{tabularx}
\end{table*}

\paragraph{Task.}
All models are trained and tested on spatial blocks of the same event, using epicentral distance as a feature. The
maps therefore describe where landslides occurred in that event. They are event-conditional susceptibility maps, not
forecasts for a future earthquake.

\paragraph{Features.}
Terrain features come from the Copernicus GLO-30 surface model \citep{esa2021dem}, computed at 30\,m and aggregated:
elevation, mean, maximum and standard deviation of slope, aspect, curvature, topographic position at 150\,m and 600\,m,
relative height, relief, and epicentral distance. GLO-30 derives from TanDEM-X data acquired between December 2010 and
January 2015. It therefore pre-dates Iburi and largely pre-dates Gorkha, but post-dates Haiti, where some landslide
scars may affect the terrain features. Observation features are distance to major roads, public roads and tracks
\citep{osm2026} and built-up density (ESA WorldCover 2021, class 50 only) \citep{zanaga2022}. Other WorldCover classes
are excluded because they show landslide scars. Roads and built-up land also post-date all three events, so post-event
reconstruction is a limitation, chiefly for the real-inventory test. In the simulations the same fields define detection,
so the mismatch does not arise. No shaking variable is used: epicentral distance is a proxy for shaking.

\begin{figure*}[t]
\centering
\includegraphics[width=0.72\textwidth]{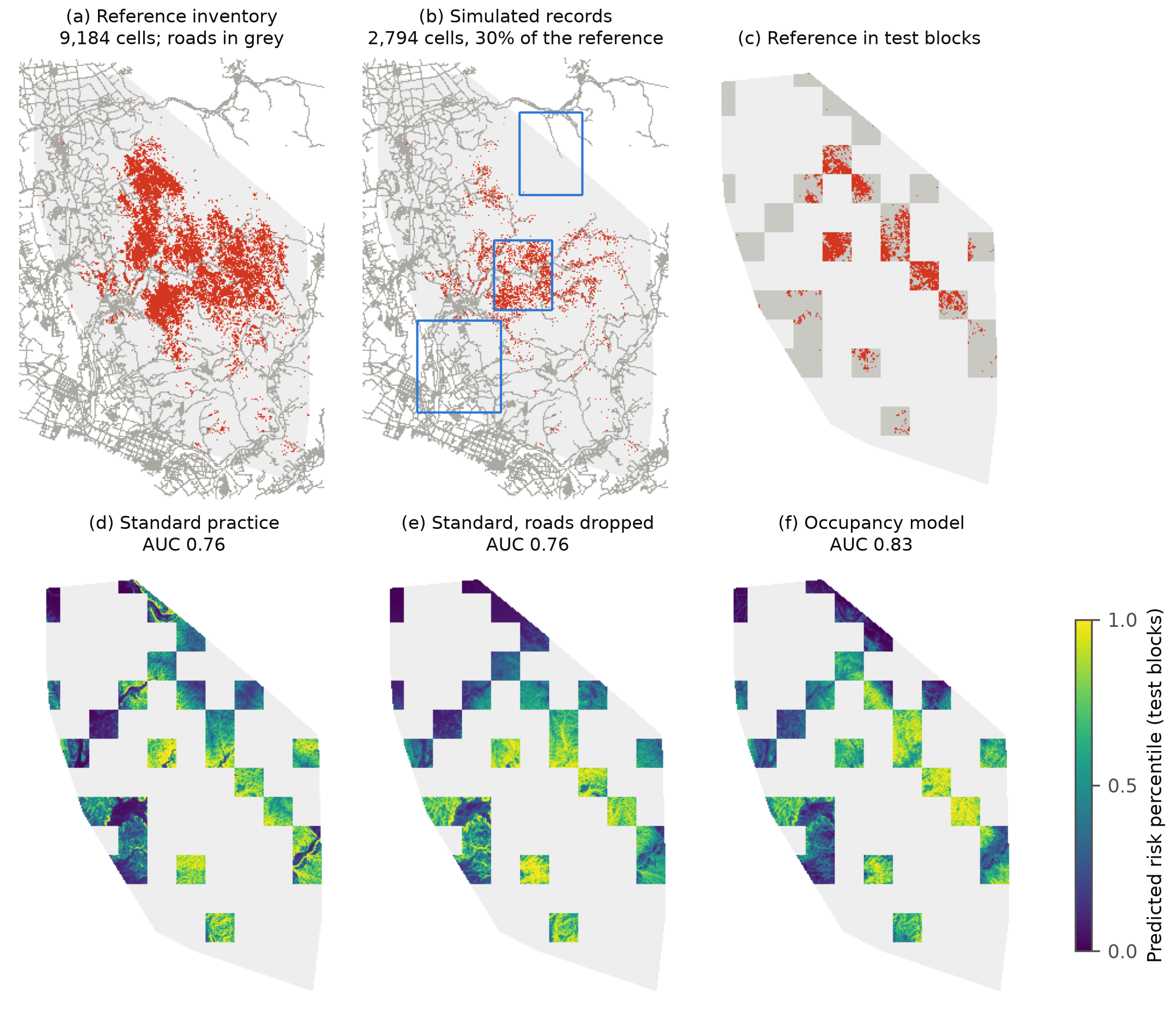}
\caption{Iburi semi-synthetic setup (scenario A, draw 0). (a) Reference inventory with public roads. (b) Simulated biased
records, with satellite footprints in blue. (c) Reference landslides in the held-out test blocks. (d--f) Predicted risk
percentile from three models. This draw is favourable to the occupancy model. Means over five draws are 0.754 for
standard practice and 0.772 for the occupancy model.}
\label{fig:maps}
\end{figure*}

\paragraph{Simulated observers.}
The reference inventory is taken as the true $z$. Three observers record it under known rules, with distance $d$ in
kilometres and standardised built-up density $u$ and cloud field $c$:
\begin{equation}
  \begin{split}
  \operatorname{logit} p_{\text{news}} &= -1 - 2 d_{\text{major}} + 0.8\,u,\\
  \operatorname{logit} p_{\text{agency}} &= 0.5 - 5 d_{\text{public}},\\
  \operatorname{logit} p_{\text{satellite}} &= -1.8\,c.
  \end{split}
\end{equation}
The satellite observer covers only three random rectangular footprints. Cloud cover is synthetic: a smoothed random
field (weight 0.6) combined with elevation (weight 0.4), redrawn for each draw and given to the detection head. Together
the observers record 18--41\% of reference landslide cells. Scenario A uses these rules. Scenario B lets susceptibility
also use road distance. Scenario C misspecifies detection with a hidden reporting-effort field, a road $\times$ built-up
interaction, and surveys that skip steep slopes. Scenario D plants a known road-cut effect, with probability
$0.25\,e^{-d/120\,\text{m}}\sigma(2\,\text{slope})$, and scenario E makes the news catalogue re-publish half of the
agency's records.

\paragraph{Real observers.}
Three real Gorkha inventories from the USGS ground-failure repository \citep{schmitt2017} are used without simulation.
Their coverage is the convex hull of each inventory within the window. The anchor area is taken from the reference
inventory, so the model sees about 2\% of the reference. Models are scored against the reference and against a consensus
(a cell in at least 2 of the 4 inventories). The consensus includes the training records, so it is not an independent
test, and it can favour models fitted to them.

\paragraph{Baselines.}
\emph{Standard practice} is one MLP trained on the union of all records, including anchor records, with terrain and
access features. \emph{Standard, roads dropped} uses terrain features only. \emph{Neutralised} models are fitted to the
same union with access features, which are then set to their mean for prediction. We implement this as an MLP, a logistic
GLM and a generalised additive model (GAM) with cubic splines. \citet{steger2021} instead use a generalised additive mixed model with
bias variables as random effects, zeroed for prediction. We follow the same logic, but with access features as fixed effects.
In the additive GLM and GAM, a constant access value gives the same ranking as zeroing those terms, but the fitted
coefficients differ because no random-effect shrinkage is applied. Our results therefore evaluate neutralisation in these
three model classes, not the mixed-model method itself. Where we refer to the \emph{best} neutralised
model, we take the highest AUC among the variants on each draw, which favours neutralisation because the best variant
cannot be known in practice. The \emph{oracle} is an MLP trained on the reference inventory.

\paragraph{Evaluation.}
Spatial blocks of $25 \times 25$ cells are split 60/15/25 into training, validation and test sets. Test-block metrics are
AUC against the reference, AUC within the most remote quarter of road distance, and road bias. With $R$ the ratio of
predicted rates in the most remote and nearest quarters of road distance, divided by the same ratio in the reference, road
bias is $|\log R|$. The oracle's road bias (0.11--0.45) indicates the noise floor of this metric. For probabilities we
report the ratio of mean predicted to true rate, expected calibration error (ECE; 10 equal-frequency bins), log loss, Brier
score and calibration slope (the coefficient from a logistic regression of outcomes on predicted log-odds; 1 is ideal, and
values below 1 mean predictions are too extreme). Each setting has five random \emph{draws} of the reporting realisation,
split and anchor. Comparisons within a scenario are paired on identical draws, using counts and Wilcoxon signed-rank tests
\citep{wilcoxon1945}. Splits differ between scenarios A and C, so we do not compare results across them. With five draws a
two-sided test cannot go below $p = 0.0625$, and draws within an earthquake share terrain and truth, so we report site-level
effects first and treat pooled $p$-values as descriptive.

\section{Results}

\subsection{Reporting bias distorts standard maps}\label{sec:bias}
Standard practice rates remote terrain far too low in every setting: median $R$ is 0.06--0.17 on Iburi and Gorkha (1 =
unbiased). On Gorkha it also loses much of its ranking skill (AUC 0.685 vs 0.857 for the oracle in scenario A).

\subsection{Occupancy model, standard practice and neutralisation}\label{sec:headline}

\begin{table*}[t]
\centering
\caption{Test AUC against the reference inventory (mean $\pm$ SD over 5 draws). Bold marks the best non-oracle model in each
column. The GAM was added after the Haiti prediction was recorded.}
\label{tab:headline}
\small
\begin{tabular}{@{}lccccc@{}}
\toprule
Model & Iburi A & Iburi C & Gorkha A & Gorkha C & Haiti A \\
\midrule
Standard practice & $0.754 \pm 0.078$ & $0.789 \pm 0.097$ & $0.685 \pm 0.088$ & $0.698 \pm 0.033$ & $0.739 \pm 0.030$ \\
Neutralised MLP & $0.736 \pm 0.029$ & $0.721 \pm 0.068$ & $0.806 \pm 0.042$ & $0.797 \pm 0.038$ & $\mathbf{0.784} \pm 0.012$ \\
Neutralised GLM & $0.713 \pm 0.035$ & $0.619 \pm 0.137$ & $\mathbf{0.820} \pm 0.028$ & $\mathbf{0.820} \pm 0.018$ & $0.783 \pm 0.015$ \\
Neutralised GAM & $0.731 \pm 0.028$ & $0.703 \pm 0.099$ & $0.814 \pm 0.018$ & $0.792 \pm 0.028$ & $0.780 \pm 0.019$ \\
Occupancy model & $\mathbf{0.772} \pm 0.060$ & $\mathbf{0.799} \pm 0.054$ & $\mathbf{0.820} \pm 0.030$ & $0.780 \pm 0.027$ & $0.773 \pm 0.023$ \\
Oracle & $0.822 \pm 0.044$ & $0.837 \pm 0.051$ & $0.857 \pm 0.007$ & $0.860 \pm 0.010$ & $0.802 \pm 0.018$ \\
\bottomrule
\end{tabular}
\end{table*}

\begin{figure*}[t]
\centering
\includegraphics[width=\textwidth]{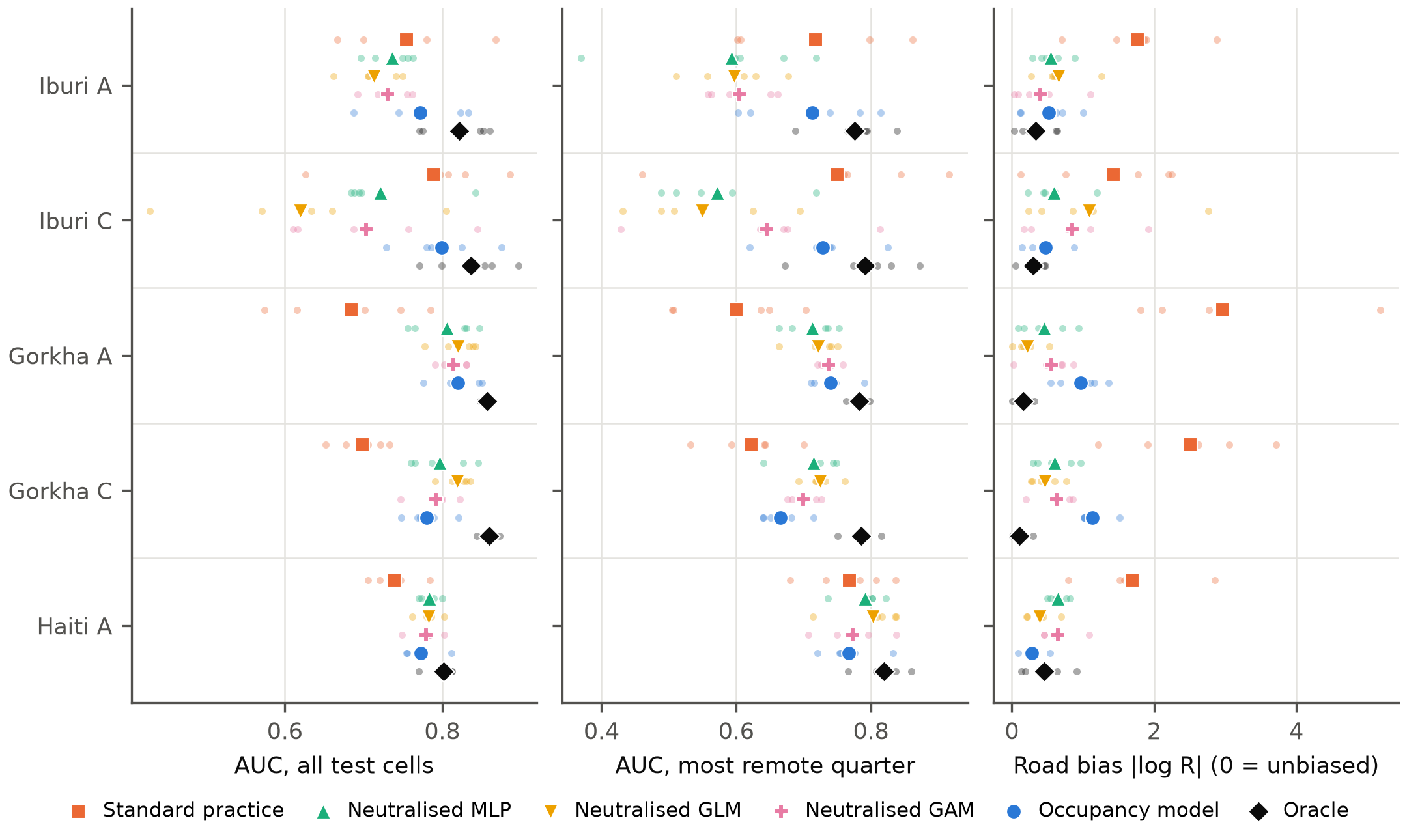}
\caption{Standard practice, three neutralised models, the occupancy model and the oracle in each setting. Large markers are
means and small dots are the 5 draws. Horizontal axes do not start at zero.}
\label{fig:headline}
\end{figure*}

\paragraph{Against standard practice.}
The occupancy model has lower road bias in 24 of 25 paired draws across all settings (Fig.~\ref{fig:headline}). Its effect on
ranking depends on the site (Table~\ref{tab:headline}). AUC gains are large on Gorkha ($+0.135$ in A, $+0.082$ in C, 5/5 draws
each) and moderate on Haiti ($+0.034$, 5/5). On Iburi they are small and unreliable ($+0.018$, 3/5 in A; $+0.010$, 2/5 in C), and
remote-quarter AUC is about equal. On Iburi, then, the benefit lies in removing bias, not in better ranking.

\paragraph{Against neutralisation.}
Neither approach dominates, and the split follows the earthquake. Relative to the best neutralised model, the occupancy model has
higher AUC on Iburi ($+0.032$, 3/5 in A; $+0.065$, 5/5 in C) and lower AUC on Gorkha ($-0.006$, 2/5 in A; $-0.045$, 0/5 in C) and
Haiti ($-0.021$, 1/5). Compared with the neutralised GAM alone, the occupancy model is higher on Iburi in 9/10 draws for AUC and 10/10
for remote-quarter AUC. On Gorkha the two are about even (AUC 6/10, mean $-0.003$; remote quarter 3/10, mean $-0.015$). Neutralisation
removes road bias as well or better: pooled over Iburi and Gorkha, the occupancy model has lower road bias than the MLP, GLM and GAM
in only 6, 7 and 7 of 20 draws. On Gorkha, road bias is higher than the GLM's in all 10 draws (0.97 and 1.13 vs 0.22 and 0.47). Haiti
is an exception, where the occupancy model has the lowest mean road bias (Table~\ref{tab:haiti}).

\subsection{Estimating landslide probability}\label{sec:calib}

\begin{table*}[t]
\centering
\caption{Probability estimates on test blocks, 25 re-fitted draws over the five settings of Table~\ref{tab:headline}. ``Scaled''
divides a baseline's prediction by the share of anchor landslide cells recorded by any inventory.}
\label{tab:calib}
\small
\begin{tabular}{@{}lcccccc@{}}
\toprule
Model & Pred.\ $\div$ true rate, & Mean & ECE & Log loss & Brier & Calib.\ \\
 & median (range) & $|\log \text{rate}|$ & & & & slope \\
\midrule
Standard practice & 0.30 (0.13--0.61) & 1.24 & 0.088 & 0.486 & 0.109 & 0.41 \\
Neutralised MLP & 0.24 (0.09--0.53) & 1.48 & 0.094 & 0.448 & 0.113 & 0.61 \\
Neutralised GLM & 0.29 (0.18--0.58) & 1.22 & 0.089 & 0.412 & 0.112 & 1.14 \\
Standard, scaled & 1.01 (0.20--2.64) & 0.60 & 0.109 & 0.816 & 0.131 & 0.15 \\
Neutralised MLP, scaled & 0.85 (0.21--2.39) & 0.61 & 0.091 & 0.604 & 0.124 & 0.41 \\
Neutralised GLM, scaled & 1.08 (0.28--4.34) & 0.70 & 0.122 & 0.852 & 0.144 & 0.61 \\
Occupancy model & \textbf{0.99 (0.44--1.89)} & \textbf{0.25} & \textbf{0.056} & \textbf{0.343} & \textbf{0.104} & 0.59 \\
Oracle & 1.00 (0.68--1.37) & 0.13 & 0.026 & 0.289 & 0.090 & 0.86 \\
\bottomrule
\end{tabular}
\end{table*}

\begin{figure*}[t]
\centering
\includegraphics[width=0.85\textwidth]{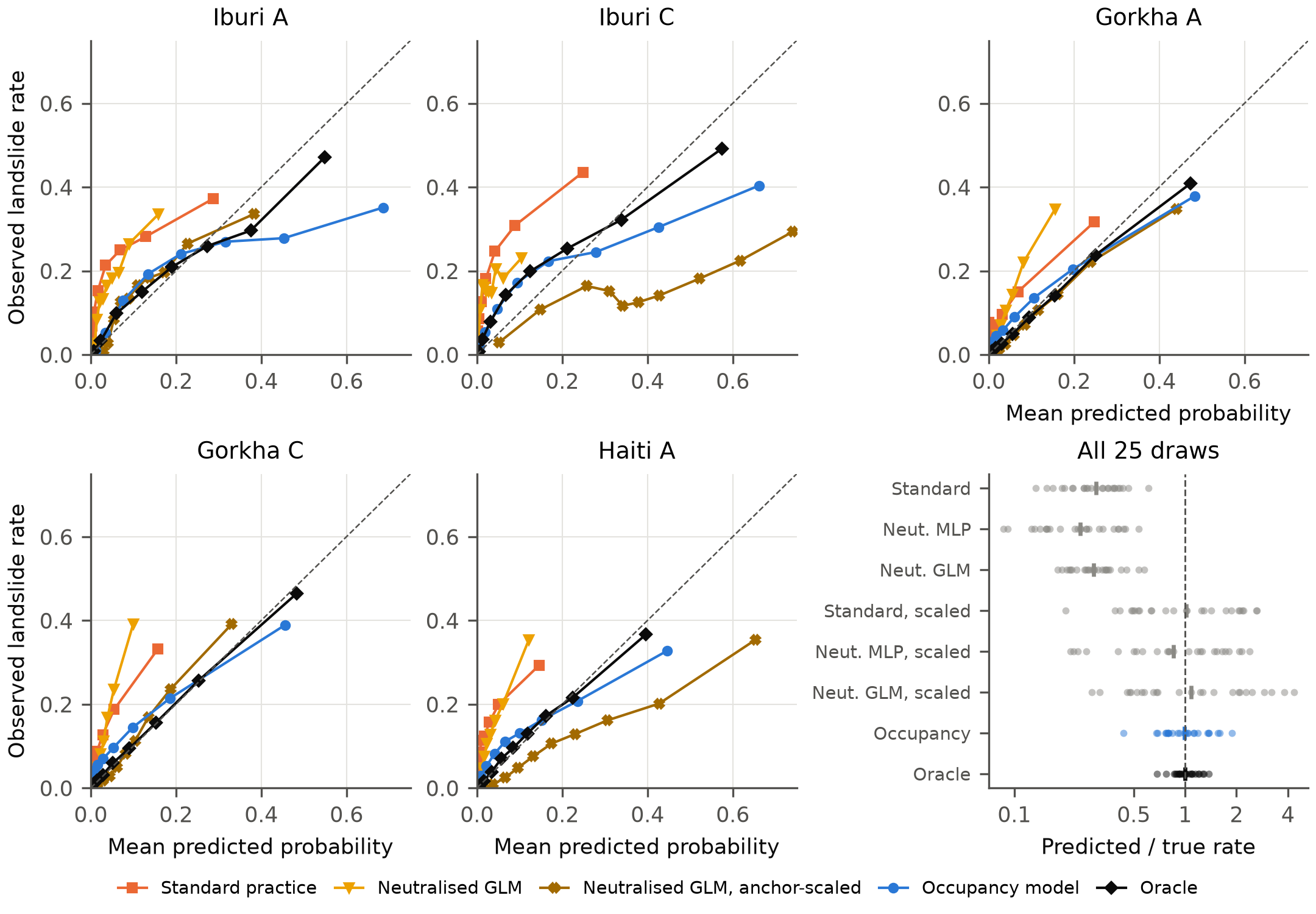}
\caption{Reliability curves on test blocks, averaged over 5 draws per setting (dashed line = perfect calibration); the neutralised
GLM stands in for the other neutralised baselines. Bottom right: predicted $\div$ true overall rate for every draw and model, on a
logarithmic axis; vertical ticks are medians.}
\label{fig:calib}
\end{figure*}

Under simulated reporting bias, baselines learn the probability of being recorded and underestimate the overall landslide rate
three- to four-fold in the median. The occupancy model's median ratio is 0.99 (Table~\ref{tab:calib}). This difference may owe more to
the anchor data than to the likelihood, so we gave the baselines the same information, dividing their predictions by the share of
anchor landslide cells recorded by any inventory \citep{elkan2008}. This corrects their median rate (0.85--1.08), so the anchor does much
of that work. The likelihood adds stability: mean $|\log \text{rate}|$ is 0.25 for the occupancy model and 0.60--0.70 for the scaled
baselines, and its rate is closer to the truth than each of theirs in 20--22 of 25 draws. Scaling also degrades other measures, because the estimate
of recording share rests on as few as 15 landslide cells and applies one factor to all cells.

Better rate estimation is not full calibration. The occupancy model has the lowest ECE, log loss and Brier score of the non-oracle
models, but its median calibration slope of 0.59 shows that its predictions are too extreme (Fig.~\ref{fig:calib}). Its rate
estimate also varies between draws (0.44--1.89) and between settings (site medians 0.80--1.19). Its probabilities should therefore
be read as better than the baselines', not as ready for direct use in risk calculations.

\subsection{When the occupancy model outperforms neutralisation}\label{sec:why}
Interaction gaps differ sharply between the earthquakes: $+0.058$ on Iburi, $0.000$ on Gorkha. To test whether this structure causes
the difference, transplant experiments keep the real terrain, roads and reporting simulation and replace only the true pattern
(Table~\ref{tab:transplant}).

\begin{table*}[t]
\centering
\caption{Interaction gaps and the AUC of the occupancy model minus the best neutralised model (MLP, GLM or GAM), 5 draws each. Upper
part: transplant worlds, with labels as in Fig.~\ref{fig:rule}. Lower part: real settings of Table~\ref{tab:headline}.}
\label{tab:transplant}
\footnotesize
\begin{tabular}{@{}llccrc@{}}
\toprule
Label & True pattern & Gap, & Gap, & Occ.\ $-$ & Occ.\ \\
 & & reference & records & best & higher \\
\midrule
I-add & Iburi terrain: additive GLM fit of the inventory & $-0.005$ & $-0.006$ & $-0.037$ & 0/5 \\
I-rand & Iburi terrain: GLM fit + synthetic interactions & $+0.056$ & $+0.037$ & $-0.019$ & 1/5 \\
I-real & Iburi terrain: flexible fit on terrain & $+0.046$ & $+0.018$ & $+0.038$ & 5/5 \\
I-real+roads & Iburi terrain: flexible fit on terrain + roads & $+0.050$ & $+0.019$ & $+0.047$ & 5/5 \\
I-no-epi & Iburi terrain: flexible fit, no epicentral distance & $+0.004$ & $-0.005$ & $-0.009$ & 2/5 \\
I-real-add & Iburi terrain: additive spline fit on terrain & $-0.006$ & $-0.010$ & $-0.018$ & 0/5 \\
G-hidden & Gorkha terrain: GLM fit + hidden road-tied driver & $+0.004$ & $-0.006$ & $-0.054$ & 2/5 \\
G-add & Gorkha terrain: additive GLM fit of the inventory & $-0.012$ & $-0.012$ & $-0.032$ & 0/5 \\
G-rand & Gorkha terrain: GLM fit + synthetic interactions & $+0.048$ & $+0.040$ & $+0.002$ & 3/5 \\
\midrule
Iburi A & Reference inventory & $+0.058$ & $+0.027$ & $+0.032$ & 3/5 \\
Iburi C & Reference inventory & $+0.055$ & $+0.047$ & $+0.065$ & 5/5 \\
Gorkha A & Reference inventory & $0.000$ & $+0.002$ & $-0.006$ & 2/5 \\
Gorkha C & Reference inventory & $+0.003$ & $-0.004$ & $-0.045$ & 0/5 \\
Haiti A & Reference inventory & $+0.002$ & $-0.014$ & $-0.021$ & 1/5 \\
\bottomrule
\end{tabular}
\end{table*}

\begin{figure*}[t]
\centering
\includegraphics[width=0.8\textwidth]{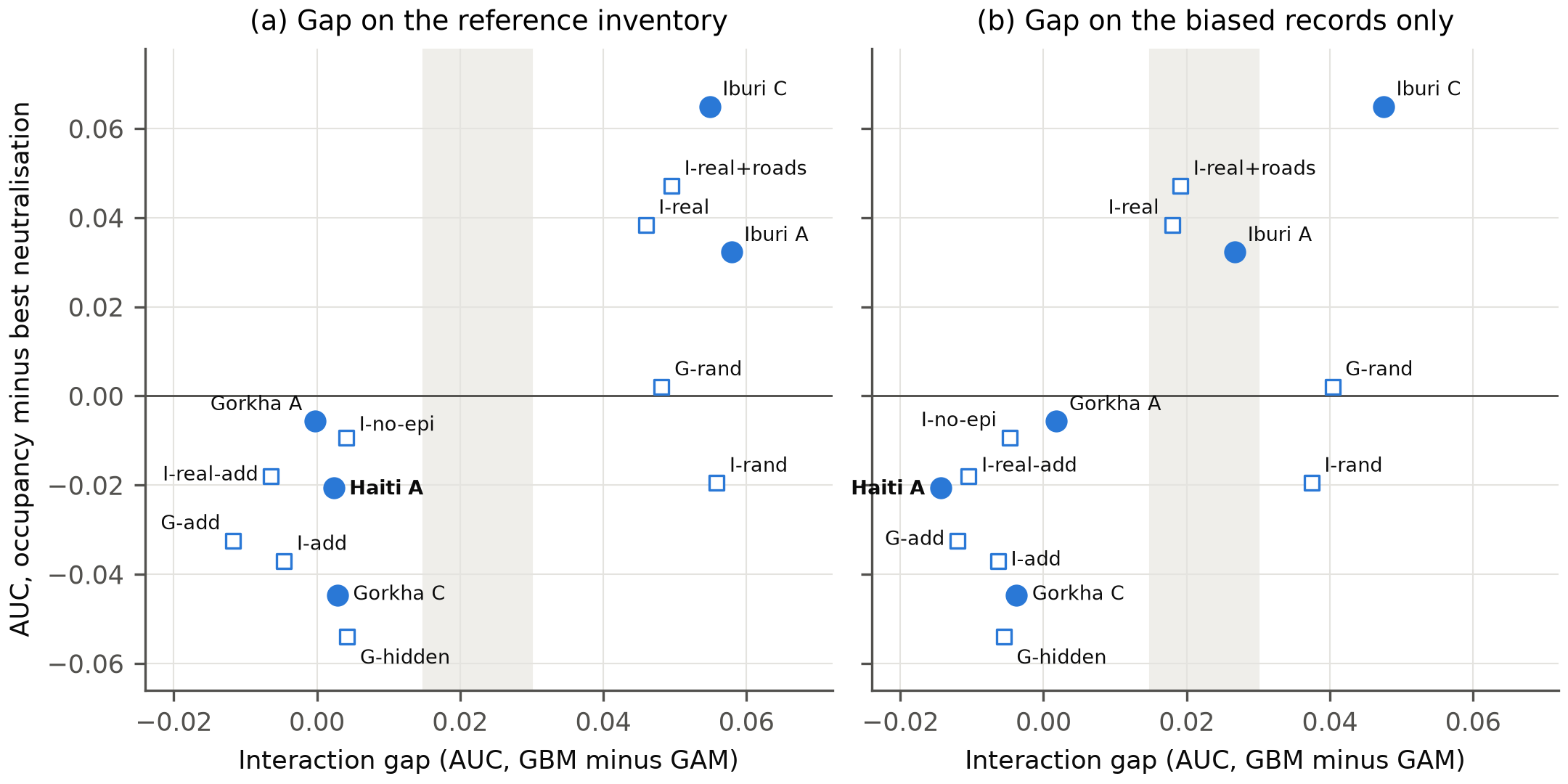}
\caption{Interaction gap against the AUC difference between the occupancy model and the best neutralised model, for real settings (filled
circles) and transplant worlds (open squares; labels as in Table~\ref{tab:transplant}). The shaded band is the no-prediction range of the
rule. Haiti, in bold, is the only setting not used in developing the rule.}
\label{fig:rule}
\end{figure*}

Three results stand out. First, Iburi's own pattern is enough to produce the advantage: refitting it from terrain alone gives $+0.038$
(5/5). Removing epicentral distance ($-0.009$, 2/5) or interactions ($-0.018$, 0/5) removes the advantage. Second, signal proxied by
roads is not the cause, since adding it barely changes the result (I-real+roads, $+0.047$), and a hand-built road-tied driver on Gorkha
terrain gives no advantage ($-0.054$, 2/5). Third, additive truths favour neutralisation on both terrains ($-0.037$ and $-0.032$), so
the outcome follows the structure of the true pattern and not the country's terrain or roads.

However, interaction strength alone is not sufficient. Synthetic interactions with a gap as large as Iburi's real pattern give no
advantage: on Iburi terrain the neutralised MLP outperforms the occupancy model (I-rand, $-0.019$, 1/5), and on Gorkha terrain the two
are tied (G-rand, $+0.002$, 3/5). The neutralised MLP retains these synthetic interactions (AUC 0.852 on I-rand) but not Iburi's real
pattern (0.776 on I-real). Something more specific to the real pattern is at work, which we discuss in Sect.~\ref{sec:discussion}.

\paragraph{The rule on biased records.}
Biased records weaken the gap (Fig.~\ref{fig:rule}b). On Iburi it falls from $+0.058$ to $+0.027$, inside the no-prediction range. Across the
14 settings of Table~\ref{tab:transplant}, a records gap below 0.015 occurs in 8 settings, and the occupancy model does not outperform the best
neutralised model in any of them. Of the 3 settings above 0.03, one shows a clear advantage (Iburi C), one a tie (G-rand) and one a loss
(I-rand). The 3 settings in the no-prediction range all favour the occupancy model. Using the reference inventory instead, the same 8 settings
fall below 0.015, and 5 of the 6 above 0.03 show a higher mean AUC for the occupancy model, one of them the tie. A low gap is therefore a useful
signal that neutralisation suffices, and is what can be claimed from records. A high gap is necessary but not sufficient. Apart from Haiti, these
settings informed the rule or derive from the same two earthquakes, so they are not independent tests.

\subsection{A prespecified test on a third earthquake (Haiti 2010)}\label{sec:haiti}
We turned the rule into a prediction for an earthquake not used in developing it. We built the Haiti grid from the \citet{harp2016} inventory and
measured the interaction gap on the reference before running any comparison: $+0.002$, below the 0.015 threshold. The prediction, that neutralisation
would be as good or better, was saved with a UTC timestamp (see code availability) but was not lodged with a third-party registry. The comparison, then
limited to the MLP and GLM variants, confirmed it. The occupancy model trailed the better variant on each draw by 0.017 AUC on average and exceeded it in
only 1 of 5 draws (differences $-0.032$, $-0.034$, $-0.016$, $+0.012$, $-0.014$). Adding the GAM afterwards gives $-0.021$, still 1/5, and the gap
computed from biased records leads to the same prediction ($-0.014$).

\begin{table}[t]
\centering
\caption{Haiti 2010, scenario A, means over 5 draws. ``Remote'' is AUC in the most remote quarter; ``Rate'' is predicted $\div$ true rate; ``Bias'' is road bias.}
\label{tab:haiti}
\small
\setlength{\tabcolsep}{5pt}
\begin{tabular}{@{}lcccc@{}}
\toprule
Model & AUC & Remote & Rate & Bias \\
\midrule
Standard practice & 0.739 & 0.768 & 0.24 & 1.70 \\
Neutralised MLP & 0.784 & 0.791 & 0.19 & 0.65 \\
Neutralised GLM & 0.783 & 0.803 & 0.29 & 0.39 \\
Neutralised GAM & 0.780 & 0.773 & 0.11 & 0.65 \\
Occupancy model & 0.773 & 0.767 & \textbf{0.97} & \textbf{0.28} \\
Oracle & 0.802 & 0.819 & 0.99 & 0.45 \\
\bottomrule
\end{tabular}
\end{table}

The other simulation findings replicate here (Table~\ref{tab:haiti}). The occupancy model outperforms standard practice in all 5 draws ($+0.034$ AUC) and has
lower road bias in all 5. Its mean rate ratio is close to 1 (0.97; median 0.80 in the re-fitted draws). Its road bias is below the oracle's, within the metric's
noise, and lower than the GLM's in 3 of 5 draws. This is a single new site that tested only the negative side of the rule. A planned fourth site (1994
Northridge) was dropped because road data for its northern, landslide-rich strip could not be completed.

\subsection{Real, independently produced inventories}\label{sec:real}
The three real Gorkha inventories are not road-biased: their catch rate of reference landslides is nearly flat with road distance (Gnyawali 45--56\% across
quartiles). They are slope-biased (Valagussa catches 28\% of gentle-slope vs 56\% of steep-slope landslides) and disagree heavily with the reference: 29--41\%
of each one's cells are absent from it, of which 59--84\% lie within one cell of a reference landslide (vs 20\% for a random cell). This suggests that many
disagreements are positional, although some may be omissions from the reference, which is itself imperfect. Together the three inventories catch 64\% of
reference cells.

\begin{table*}[t]
\centering
\caption{Real Gorkha inventories (5 draws, held-out blocks). Slope bias is $|\log|$ of the steep/gentle predicted-rate ratio relative to the reference. The
consensus includes the training records, so it is not an independent test.}
\label{tab:real}
\small
\begin{tabularx}{\textwidth}{@{}Lcccc@{}}
\toprule
Model & AUC vs reference & AUC vs consensus & Slope bias & Pred.\ $\div$ true rate \\
\midrule
Standard practice & 0.830 & 0.867 & 0.44 & 1.11 \\
Standard, roads dropped & 0.836 & 0.872 & 0.41 & 1.13 \\
Neutralised MLP & 0.829 & 0.863 & \textbf{0.25} & 1.15 \\
Neutralised GLM & 0.825 & 0.858 & 0.28 & \textbf{1.05} \\
Occupancy model & \textbf{0.837} & \textbf{0.873} & 0.47 & 1.38 \\
Occupancy + visibility detection & 0.835 & 0.872 & 0.31 & 1.35 \\
Occupancy + visibility + constant false positives & \textbf{0.837} & \textbf{0.873} & 0.86 & 1.15 \\
Oracle & 0.852 & 0.867 & 0.24 & 0.99 \\
\bottomrule
\end{tabularx}
\end{table*}

Here the occupancy model brings no meaningful gain (Table~\ref{tab:real}). Its AUC ties the terrain-only standard model ($+0.001$, 3/5 draws, $p = 0.44$) and is $0.008$
and $0.012$ above the neutralised MLP and GLM (4/5 and 5/5). These differences are small. The records gap here is $0.000$, so the rule predicts that neutralisation will
be as good or better; on AUC it was narrowly not. The occupancy model has the worst slope bias of the main models
(0.47) and overestimates the rate by 38\%, while neutralisation gives a lower slope bias and a rate near 1. Adding visibility features to detection improves slope bias
but does not fix the rate. Adding constant false positives improves the rate (1.15) but nearly doubles slope bias (0.86). The overestimated rate is consistent with
assumption (iii): every record is treated as a true landslide, so cells absent from the reference inflate $\psi$. The model does rank the completeness of each inventory
correctly with access to only the anchor part of the reference (learned 0.59--0.65, 0.48--0.56 and 0.11--0.13 vs measured 0.52, 0.43 and 0.10), but overestimates each.

Because most extra cells sit next to a reference landslide, we tried the spillover term (Eq.~\ref{eq:spill}). It was not identifiable: the fit swapped freely between ``recorded
because a landslide is here'' and ``recorded because one is next door''. In one draw almost every record was attributed to spillover (detection 0.01--0.08; AUC 0.731 vs 0.858
without the term). In another the opposite occurred (detection 0.21--0.85; AUC 0.820 vs 0.821; rate underestimated about threefold). We stopped after these two draws. Correcting
real false positives probably needs outside information, such as a small hand-checked sample \citep{miller2011}.

\subsection{Robustness and design choices}\label{sec:robust}

\paragraph{Monotone detection.}
On Iburi, monotone detection raised remote-quarter AUC from 0.58 to 0.72, and it is the design choice with the largest effect. A spatial random effect in detection made results
worse ($-0.01$ to $-0.02$ AUC), probably because it competes with the spatial pattern of susceptibility.

\begin{figure*}[t]
\centering
\includegraphics[width=0.8\textwidth]{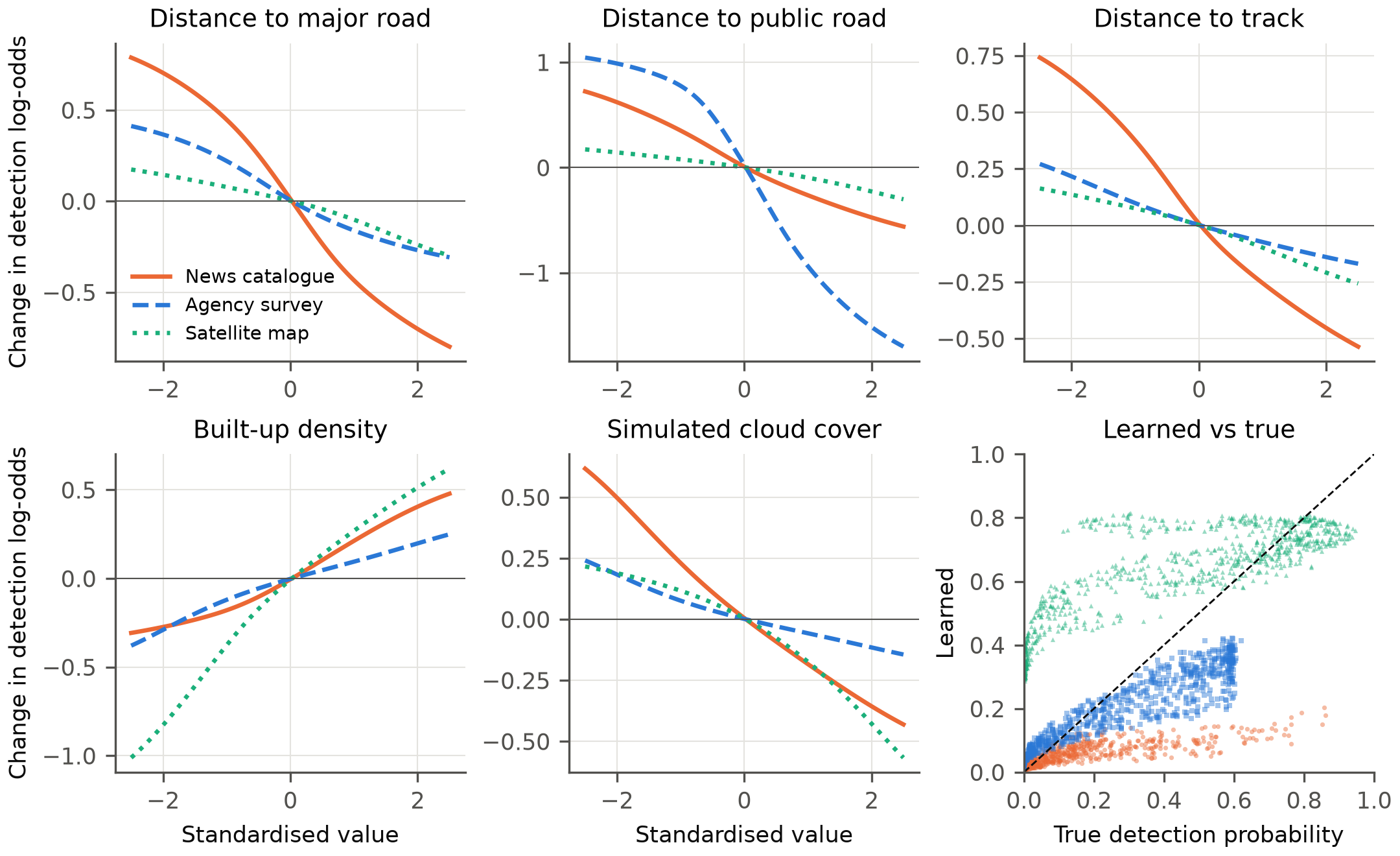}
\caption{Detection learned for each simulated observer (Iburi, scenario A, draw 0). Curves show the additive effect of each feature on detection log-odds, constrained to the
known direction. The last panel compares learned and true detection probability (dashed line = equality; colours as in the other panels).}
\label{fig:curves}
\end{figure*}

\paragraph{What the model learns about observers.}
Learned detection follows the direction of the main effects (Fig.~\ref{fig:curves}), with correlations of 0.81 (news), 0.80 (agency) and 0.56 (satellite) between learned and true
probability. Magnitudes are recovered poorly: the news catalogue's detection is strongly underestimated and the satellite's overestimated. Since the overall rate is still near
correct, some errors may offset one another, and the separation of $\psi$ and $p$ should not be over-interpreted. The satellite observer also picks up spurious road and built-up
effects.

\paragraph{Misspecification and dependence.}
Under misspecified detection (scenario C) the model keeps its advantage on Iburi but falls behind the neutralised GLM on Gorkha (0.780 vs 0.820). When the news catalogue copies half of
the agency's records (scenario E, on the same draws as A), Iburi AUC moves only from 0.772 to 0.776.

\paragraph{Anchor alone.}
Models trained only on the anchor area give the worst maps on Iburi (AUC 0.68--0.70), so the anchor is useful within the model but insufficient by itself.

\paragraph{Roads that genuinely cause landslides.}
Letting road distance enter susceptibility freely brings the bias back (Iburi B: AUC 0.730, median $R = 0.33$). Restricting it to the short-range term (Eq.~\ref{eq:roadcut}) recovers
performance (AUC 0.774, road bias below standard practice in 5/5 draws), about the same as leaving roads out entirely (0.772). With a planted road-cut effect (scenario D), the short-range
model reproduces the near-road excess most closely (lift $1.02 \pm 0.04$, where 1 matches the reference; free roads 1.27; standard practice 1.48). The learned strength responds only
modestly (0.33 to 0.43 log-odds), and the range is over-estimated (185 vs 120\,m). The term thus allows a road effect without re-absorbing bias, but does not measure that effect well.

\section{Discussion}\label{sec:discussion}

\paragraph{Which method to use.}
Both approaches remove much of the road-driven reporting bias, but they have different costs and strengths. Neutralisation needs one inventory and, where susceptibility is close to additive, it ranks
at least as well. The occupancy model needs at least two inventories with different biases and a small exhaustively mapped area. In return it estimates the overall rate more stably and, in some settings,
ranks better. We suggest the following procedure. Fit a gradient-boosting and an additive model to the available records and compute the interaction gap on held-out blocks. If the gap is below 0.015,
use neutralisation for ranking, since nothing in our tests favoured the extra data requirement. If it is larger, the occupancy model may help, but this is not assured. In either case, if the overall
rate matters and an anchor area exists, the occupancy model gave more stable rates than scaling a neutralised model.

\paragraph{Mechanism.}
Neither generic nonlinearity nor road-proxied signal explains the Iburi advantage; the combination of epicentral structure and interactions in the real pattern does. One reading is that a flexible
neutralised model fits part of that structure through the road variables, and discards it when they are neutralised. Slope is more strongly correlated with road distance on Iburi ($r = 0.50$--$0.59$ for
public and major roads) than on Gorkha ($0.39$) or Haiti ($0.23$--$0.37$), which is consistent with this reading. It is not established, however: terrain as a whole predicts road distance better on
Gorkha than on Iburi ($R^2$ 0.74 vs 0.49), and a GLM loses on Iburi for a simpler reason, as it cannot represent interactions at all. We avoid describing this as a shaking $\times$ terrain interaction, since we measured epicentral distance and not shaking. Testing the role of shaking directly, for example with USGS ShakeMap, is a natural next step.

\paragraph{Real inventories and the target of the method.}
The method was designed for reporting driven by roads and population, and the real inventories we could test have a different kind of bias. This fits earlier observations that road effects depend on how an
inventory is made \citep{malamud2022}: satellite mapping of a whole event has no need to follow roads. Slope bias is harder, because slope also drives real occurrence, which undermines the exclusion
restriction. On such inventories, a standard terrain-only model is already within 0.016 AUC of the oracle, leaving little room for improvement. Scoring against a near-complete reference also matters for all these comparisons: \citet{steger2021} note that performance measured on biased test data can understate a bias-adjusted model. Our results do not show that the method is useful on real
inventories, only that it does not fail on ranking; for rate estimation and slope bias it was worse than the alternatives.

\paragraph{Limitations.}
\begin{itemize}
  \item Road and population reporting bias, and cloud cover, are simulated. Detection features match the simulated rules closely in scenario A, which favours the occupancy model; scenario C relaxes this.
  \item Three earthquakes, of which only Haiti was out of sample, and it tested only the negative side of the rule. There is no out-of-sample test of a high gap.
  \item Draws within one earthquake share terrain and truth, so pooled statistics overstate generality. No transplant uses Gorkha's own refitted pattern.
  \item The reference inventories are treated as truth, but are themselves imperfect.
  \item Neutralisation is implemented with fixed effects. A random-effects version, as in \citet{steger2021}, was not tested and could perform differently.
  \item Maps are conditional on the event and are not forecasts. Roads and built-up land post-date the events, and the DEM post-dates Haiti.
  \item The method needs more input than neutralisation. Exhaustively mapped areas may not exist, and the scaled-baseline comparison used one simple correction; other corrections may do better.
  \item Cells are square at 120--210\,m and features are tabular. Slope units and image-based features are natural extensions, since the framework accepts any susceptibility network.
\end{itemize}

\paragraph{Main open test.}
The bias the method targets is typical of report-based catalogues (news, agency and citizen reports). NASA's Global Landslide Catalog \citep{kirschbaum2010,kirschbaum2015} cannot be scored against the
co-seismic inventories used here because it covers rainfall-triggered landslides only. A decisive test needs a rainfall event that has both report-based records and a complete inventory, for example
Hurricane Maria in Puerto Rico in 2017 \citep{bessettekirton2019}.

\section{Conclusions}
Treating several biased landslide inventories as imperfect observers reduces simulated road and population bias relative to standard practice and gives more stable estimates of the overall landslide rate than
baselines, including baselines rescaled with the same anchor area, although cell-level probabilities remain too extreme. Whether it also ranks better than neutralisation depends on the structure of the true
landslide pattern: it did on Iburi, but not on Gorkha or Haiti. Wherever the interaction gap computed from biased records alone was low, neutralisation was at least as accurate, including a
prediction recorded before testing on Haiti, whereas a high gap did not guarantee an advantage. On real satellite-mapped inventories, whose bias follows slope and not roads, the model brought no meaningful gain.
Neutralisation remains the sensible default for ranking when only one inventory exists or the interaction gap is low. The case for the occupancy model rests on rate estimation and on patterns with strong
interactions, and its value on the report-based catalogues it was designed for is still to be shown.

\section*{Data and Code Availability}
Inventories: Zenodo record 2577300 (Iburi) \citep{zhang2019}; USGS data release \url{https://doi.org/10.5066/F7DZ06F9} (Gorkha reference) \citep{zekkos2017}; USGS ground-failure repository
\citep{schmitt2017} for the real Gorkha inventories and the Haiti inventory \citep{harp2016}. DEM: Copernicus GLO-30 \citep{esa2021dem}. Land cover: ESA WorldCover 2021 \citep{zanaga2022}. Roads: OpenStreetMap
via Overpass \citep{osm2026}. Code, results and the timestamped Haiti prediction: \url{https://github.com/rashiedomar/occupancy-landslide-inventories}.

\bibliographystyle{unsrtnat}
\bibliography{refs}

\end{document}